\documentclass[letterpaper]{article} 
\usepackage{aaai24}  
\usepackage{times}  
\usepackage{helvet}  
\usepackage{courier}  
\usepackage[hyphens]{url}  
\usepackage{graphicx} 
\def\UrlFont{\rm}  
\usepackage{natbib}  
\usepackage{caption} 
\usepackage{algorithm}
\usepackage{algorithmic}

\usepackage{newfloat}
\usepackage{listings}
\DeclareCaptionStyle{ruled}{labelfont=normalfont,labelsep=colon,strut=off} 
\floatstyle{ruled}
\newfloat{listing}{tb}{lst}{}
\floatname{listing}{Listing}
\usepackage{amssymb}
\usepackage{amsmath}
\DeclareMathOperator*{\argmax}{arg\,max}
\DeclareMathOperator*{\argmin}{arg\,min}
\usepackage{subfigure}

\title{Virtual Action Actor-Critic Framework for Exploration\\(Student Abstract)}
\author {
    Bumgeun Park,
    Taeyoung Kim,
    Quoc-Vinh Lai-Dang,
    Dongsoo Har
}
\affiliations {
    Korea Advanced Institute of Science and Technology, Daejeon 34051, Korea\\
    j4t123@kaist.ac.kr, ngng9957@kaist.ac.kr, ldqvinh@kaist.ac.kr, dshar@kaist.ac.kr
}

\begin{document}

\maketitle

\begin{abstract}
Efficient exploration for an agent is challenging in reinforcement learning (RL). In this paper, a novel actor-critic framework namely virtual action actor-critic (VAAC), is proposed to address the challenge of efficient exploration in RL. This work is inspired by humans' ability to imagine the potential outcomes of their actions without actually taking them. In order to emulate this ability, VAAC introduces a new actor called virtual actor (VA), alongside the conventional actor-critic framework. Unlike the conventional actor, the VA takes the virtual action to anticipate the next state without interacting with the environment. With the virtual policy following a Gaussian distribution, the VA is trained to maximize the anticipated novelty of the subsequent state resulting from a virtual action. If any next state resulting from available actions does not exhibit high anticipated novelty, training the VA leads to an increase in the virtual policy entropy. Hence, high virtual policy entropy represents that there is no room for exploration. The proposed VAAC aims to maximize a modified Q function, which combines cumulative rewards and the negative sum of virtual policy entropy. Experimental results show that the VAAC improves the exploration performance compared to existing algorithms.
\end{abstract}

\section{Introduction}
Reinforcement learning (RL) enables an agent to maximize the expected reward. When state-action dimension is high or the reward is sparse, it is hard for the agent to determine the action maximizing the expected reward. For the high performance of the agent, it is crucial to explore various situations. To address this issue, many RL algorithms have employed diverse methods such as incorporating policy entropy into the training objective to encourage more uniform action selection or employing intrinsic rewards that are the bonuses assigned to unfamiliar states. A straightforward solution for better exploration is to assign higher intrinsic rewards to the unfamiliar states than to the frequently visited ones. In order to maximize the reward, including intrinsic rewards, the policy is trained to guide the agent towards unfamiliar states. Paradoxically, this explicitly requires prior experience in unfamiliar states. In this paper, we introduce a novel method to enhance exploration that does not require prior experiences in unfamiliar states.

\section{Method}
Humans often think about the potential outcomes of their actions and anticipate how different those outcomes might be from their experience. This study aims to replicate these capabilities in artificial agents. These capabilities involve predicting the next state in a given state-action pair and measuring how novel and unfamiliar the predicted state is. In order to achieve this, we introduce the virtual action actor-critic (VAAC) framework utilizing the proposed virtual actor (VA) with the conventional off-policy actor-critic framework and propose an anticipated novelty reward module (ANRM) to train VA.

The ANRM utilizes dynamic model $p:\mathcal{S}\times\mathcal{A}\rightarrow\mathcal{S}$ to predict the next state \cite{dynamic_model} and the novelty module $f:\mathcal{S}\rightarrow\mathbb{R}$ \cite{rnd} to measure the novelty of states, where $\mathcal{S}$ and $\mathcal{A}$ are the state space and the action space. The ANRM can predict the novelty of the next state in a given state-action pair and it can be represented by $f \circ p:\mathcal{S}\times\mathcal{A}\rightarrow\mathbb{R}$ and corresponding novelty can be expressed as $\phi(s,a)$. The most important property of the ANRM is to predict the novelty of next states in a given state-action pair without agent-environment interaction.

The VA takes virtual action in a given state, aiming to maximize the novelty that comes from the ANRM. Its policy (virtual policy) is restricted to a Gaussian distribution as proposed by \cite{sac}. Hence, the virtual policy is updated towards the exponential of $\phi$ as given by
\begin{equation}
    \psi_{new}=\argmin_{\psi'\in\Pi}{D_{KL}\Biggl(\psi'(\cdot|s_t)\bigg\|\frac{exp\big(\phi(s_t,\cdot)\big)}{Z(s_t)}\Biggl)}\label{KL}
\end{equation}
where $\psi$, $\Pi$ and $Z$ are the virtual policy, a parameterized family of Gaussian distribution and the partition function to normalize the distribution, respectively. Since the KL-divergence in Eq.\eqref{KL} is equal to $-(\mathbb{E}_{a\sim\psi}[\phi(s_t,a_t)]+\mathcal{H}(\psi(\cdot|s_t)))$, Eq.\eqref{KL} is converted to $\psi_{new}=\argmax_{\psi'\in\Pi}{\mathbb{E}_{a_t\sim\psi}[\phi(s_t,a_t)]+\mathcal{H}(\psi(\cdot|s_t))}$. If $\phi(s_t,a_t)$ remains consistent across the different actions, updating virtual policy increases the virtual policy entropy $\mathcal{H}(\psi(\cdot|s_t))$ because $\mathbb{E}_{a_t\sim\psi}[\phi(s_t,a_t)]$ does not depend on the virtual policy $\phi$ any more. High virtual policy entropy represents a low potential for exploration because the change of action in the high $\mathcal{H}(\psi(\cdot|s_t))$ state does not directly improve the novelty in the next state. 

The proposed VAAC aims to maximize the return while minimizing the sum of virtual policy entropy as given as
\begin{equation}
    J(\pi)=\sum_{t=0}^{T}\mathbb{E}_{(s_t,a_t)\sim\rho_\pi}[r(s_t,a_t)-\mathcal{H}\big(\psi(\cdot|s_t)\big)]\label{objective}
\end{equation}
where $\rho_\pi$ is the trajectory distribution induced by $\pi$. In order to achieve this, the VAAC utilizes the virtual policy entropy in the Bellman backup operator $\mathcal{T}^{\pi}$ as given by 
\begin{equation}
    \mathcal{T}^{\pi}Q(s_t,a_t)\triangleq r(s_t,a_t)+\gamma\mathbb{E}_{s_{t+1}\sim p}[V(s_{t+1})]\label{bellman}
\end{equation}
where
\begin{equation}
    V(s_{t})=\mathbb{E}_{a_t\sim\pi}[Q(s_t,a_t)]+\mathbb{E}_{a_t\sim\psi}[\log{\psi(a_t|s_t)}]\label{bellman}
\end{equation}
and $p$ are the value function and the transition probability.

In the policy improvement step, we update the policy in the same way with \cite{sac} as given by
\begin{equation}
    \pi_{new}=\argmin_{\pi'\in\Pi}{D_{KL}\Biggl(\pi'(\cdot|s_t)\bigg\|\frac{exp\big(Q^{\pi_{old}}(s_t,\cdot)\big)}{Z^{\pi_{old}}(s_t)}\Biggl)}\label{policy}
\end{equation}
where $Z^{\pi_{old}}(s_t)$ is the partition function to normalize the distribution.

\begin{figure}[t]
\centering
\subfigure[]{\includegraphics[width=0.495\columnwidth]{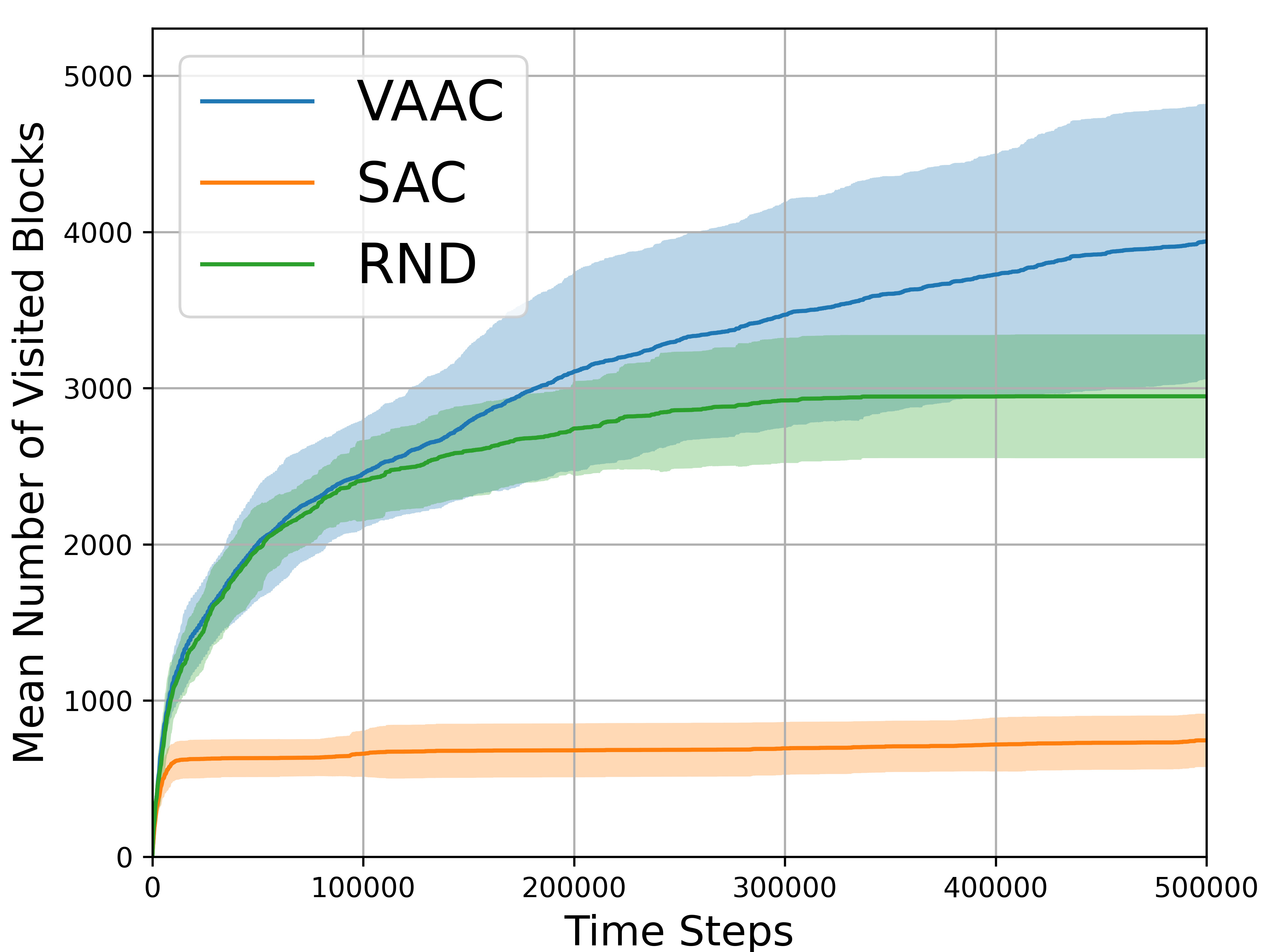}\label{a}}
\subfigure[]{\includegraphics[width=0.495\columnwidth]{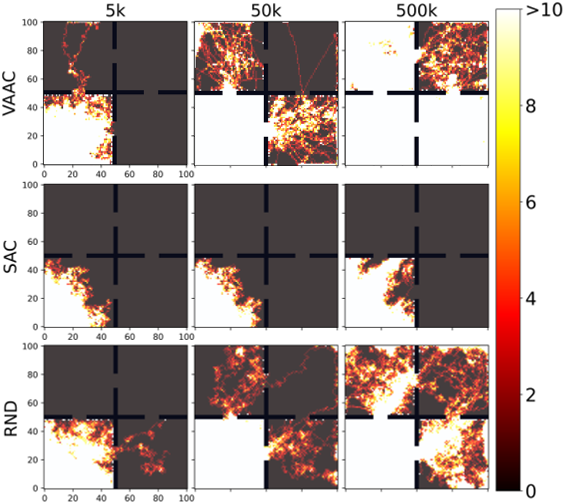}\label{b}}
\caption{Experimental results in Continuous 4-room maze: (a) Each curve represents the average number of different state visits with 30 different random seeds. The width of the shaded area represents one standard deviation $(1\sigma)$ from the mean; (b) Visit histogram obtained during 5k, 50k and 500k steps. The brighter colors indicate higher visit frequencies.}
\label{fig1}
\end{figure}

\section{Experiments}
\subsection{Experimental Setup}
We evaluate the performance of the proposed VAAC compared with soft actor-critic (SAC) and random network distillation (RND), which are the state-of-the-art algorithms for exploration, in the Continuous 4-room maze and SparseMujoco tasks used by \cite{minmax}. SparseMujoco is widely used as difficult tasks and the agent get rewards only in a certain condition. Continuous 4-room maze consists of $100\times100$ size of space with a wall and the agent can move started from $(1,1)$. The state is the $(x,y)$ position of the agent and the action is $(dx,dy)$ bounded by $[-1,1]\times[-1,1]$. After 1000 steps, the agent restarts from its initial state. To better demonstrate the pure exploration performance, Continuous 4-room maze is employed without any rewards. Due to the limited space, more detailed setup is presented in the supplementary file.

\subsection{Results}
In order to evaluate the pure exploration performance of VAAC, the training is performed during 500k steps in the Continuous 4-room maze task. The number of visits is counted by partitioned $1\times1$ squares every 1000 steps. As shown in Figure \ref{fig1}, the VAAC visited many more states than the other methods, which shows the superior pure exploration performance of the VAAC. We also conduct experiments in SparseMujoco tasks where the agent requires much exploration. Table \ref{table1} shows that the VAAC achieves the highest max average return performance across almost all tasks.

\begin{table}[t]
\centering
\resizebox{1.0\columnwidth}{!}{
\begin{tabular}{l|c|c|c}
    \hline & VAAC & SAC & RND \\ \hline
    S.HalfCheetah & $\textbf{951.6}$$\pm$$\textbf{2.5}$ & 916.3$\pm$37.8 &266.4$\pm$326.2\\ 
    S.Hopper & $\textbf{894.6}$$\pm$$\textbf{5.2}$ & 866.7$\pm$14.5&113.2$\pm$43.0 \\
    S.Walker2d & $\textbf{889.5}$$\pm$$\textbf{11.9}$ &873.6$\pm$42.1 &6.8$\pm$6.0 \\
    S.Ant &$\textbf{665.7}$$\pm$$\textbf{78.3}$ & 584.3$\pm$103.5 & $\textbf{670.8}$$\pm$$\textbf{149.9}$ \\ \hline
\end{tabular}
}
\caption{Maximum of average return by VAAC, SAC and RND in four SparseMujoco tasks after training for 1M steps.}
\label{table1}
\end{table}

\section{Discussion}
Since the performance of the proposed VAAC depends on the ANRM and VA, additional studies for the reliability of these remain as future work. Due to the complexity of the framework, training VAAC is computationally intensive, and the convergence is not theoretically guaranteed. Despite the mentioned limitations, VAAC shows good performance compared with other methods, especially in tasks where the agent needs to explore extensively.

\section{Acknowledgments}
This work was supported by the Institute for Information communications Technology Promotion (IITP) grant funded by the Korean government (MSIT) (No. 2020-0-00440, Development of Artificial Intelligence Technology that continuously improves itself as the situation changes in the real world).

\bigskip
\bibliography{aaai24}

\begin{thebibliography}{4}
\providecommand{\natexlab}[1]{#1}

\bibitem[{Burda et~al.(2018)Burda, Edwards, Storkey, and Klimov}]{rnd}
Burda, Y.; Edwards, H.; Storkey, A.; and Klimov, O. 2018.
\newblock Exploration by random network distillation.
\newblock In \emph{International Conference on Learning Representations}.

\bibitem[{Haarnoja et~al.(2018)Haarnoja, Zhou, Hartikainen, Tucker, Ha, Tan,
  Kumar, Zhu, Gupta, Abbeel et~al.}]{sac}
Haarnoja, T.; Zhou, A.; Hartikainen, K.; Tucker, G.; Ha, S.; Tan, J.; Kumar,
  V.; Zhu, H.; Gupta, A.; Abbeel, P.; et~al. 2018.
\newblock Soft actor-critic algorithms and applications.
\newblock \emph{arXiv preprint arXiv:1812.05905}.

\bibitem[{Han and Sung(2021)}]{minmax}
Han, S.; and Sung, Y. 2021.
\newblock A max-min entropy framework for reinforcement learning.
\newblock \emph{Advances in Neural Information Processing Systems}, 34:
  25732--25745.

\bibitem[{Moerland et~al.(2023)Moerland, Broekens, Plaat, Jonker
  et~al.}]{dynamic_model}
Moerland, T.~M.; Broekens, J.; Plaat, A.; Jonker, C.~M.; et~al. 2023.
\newblock Model-based reinforcement learning: A survey.
\newblock \emph{Foundations and Trends{\textregistered} in Machine Learning},
  16(1): 1--118.

\end{thebibliography}

\end{document}